%% file: paper.tex
\documentclass[sigconf]{acmart}
\usepackage{booktabs} 
\usepackage{amsmath}%
\usepackage{xspace}
\usepackage[lined, ruled, boxed, commentsnumbered,noend]{algorithm2e}
\usepackage{enumitem}

\usepackage{amsthm}
\usepackage[disable]{todonotes}
\newtheorem{informal}{Informal Problem}
\newtheorem*{formal}{Definition: Open Attribute Value Extraction}
\newcommand{\sysname}{{\sf OpenTag}\xspace}
\DeclareMathOperator{\softmax}{softmax}
\DeclareMathOperator{\argmax}{argmax}

\newcommand\blfootnote[1]{%
	\begingroup
	\renewcommand\thefootnote{}\footnote{#1}%
	\addtocounter{footnote}{-1}%
	\endgroup
}

\begin{document}
\title{OpenTag: Open Attribute Value Extraction from Product Profiles}

\author{Guineng Zheng$^\mathsection$$^*$, Subhabrata Mukherjee$^\text{\textdagger}$, Xin Luna Dong$^\text{\textdagger}$, Feifei Li$^\mathsection$}
\affiliation{%
\institution{$^\mathsection$University of Utah\quad $^\text{\textdagger}$Amazon.com\\\{guineng, 
lifeifei\}@cs.utah.edu,\quad \{subhomj, 
lunadong\}@amazon.com}
\vspace*{3em}
}





\input{abstract}

%
%



\copyrightyear{2018} 
\acmYear{2018} 
\setcopyright{acmcopyright}
\acmConference[KDD '18]{The 24th ACM SIGKDD International Conference on Knowledge Discovery \& Data Mining}{August 19--23, 2018}{London, United Kingdom}
\acmBooktitle{KDD '18: The 24th ACM SIGKDD International Conference on Knowledge Discovery \& Data Mining, August 19--23, 2018, London, United Kingdom}
\acmPrice{15.00}
\acmDOI{10.1145/3219819.3219839}
\acmISBN{978-1-4503-5552-0/18/08}

\maketitle

\input{introduction}

\input{method}

\input{base}
\input{active}
\input{experiment}
\input{related}
\input{conclusion}

\bibliographystyle{ACM-Reference-Format}
\bibliography{bibliography}

\end{document}

%% file: abstract.tex
\begin{abstract}
  Extraction of missing attribute values is to find values describing
  an attribute of interest from a free text input. Most past related
  work on extraction of missing attribute values work with a closed
  world assumption with the possible set of values known
  beforehand, or use dictionaries of values and hand-crafted features. How can we discover new attribute values that we have
  never seen before? Can we do this with limited human annotation or
  supervision? We study this problem in the context of product
  catalogs that often have missing values for many attributes of
  interest.

  In this work, we leverage product profile information such as titles
  and descriptions to discover missing values of product
  attributes. We develop a novel deep tagging model OpenTag for this
  extraction problem with the following contributions: (1) we
  formalize the problem as a sequence tagging task, and propose a
  joint model exploiting recurrent neural networks (specifically,
  bidirectional LSTM) to capture context and semantics, and
  Conditional Random Fields (CRF) to enforce tagging consistency; (2)
  we develop a novel attention mechanism to provide interpretable
  explanation for our model's decisions; (3) we propose a novel
  sampling strategy exploring active learning to reduce the burden of
  human annotation. OpenTag does not use any dictionary or hand-crafted features as in prior works. Extensive experiments in real-life datasets in
  different domains show that OpenTag with our active learning strategy discovers new attribute
  values from as few as $150$ annotated samples (reduction in $3.3$x
  amount of annotation effort) with a high F-score of $83\%$,
  outperforming state-of-the-art models.
  \blfootnote{$^*$Work performed during internship at Amazon.}
\end{abstract}

%% file: introduction.tex
\section{Introduction}\label{sec:introduction}
Product catalogs are a valuable resource for eCommerce retailers that
allow them to organize, standardize, and publish information to
customers. However, this catalog information is often noisy and
incomplete with a lot of missing values for product attributes. An
interesting and important challenge is to supplement the catalog with
missing values for attributes of interest from
product descriptions and other related product information,
especially with values that we have never seen before.

\begin{figure}[!t]
	\begin{center}
		\includegraphics[width=\linewidth]{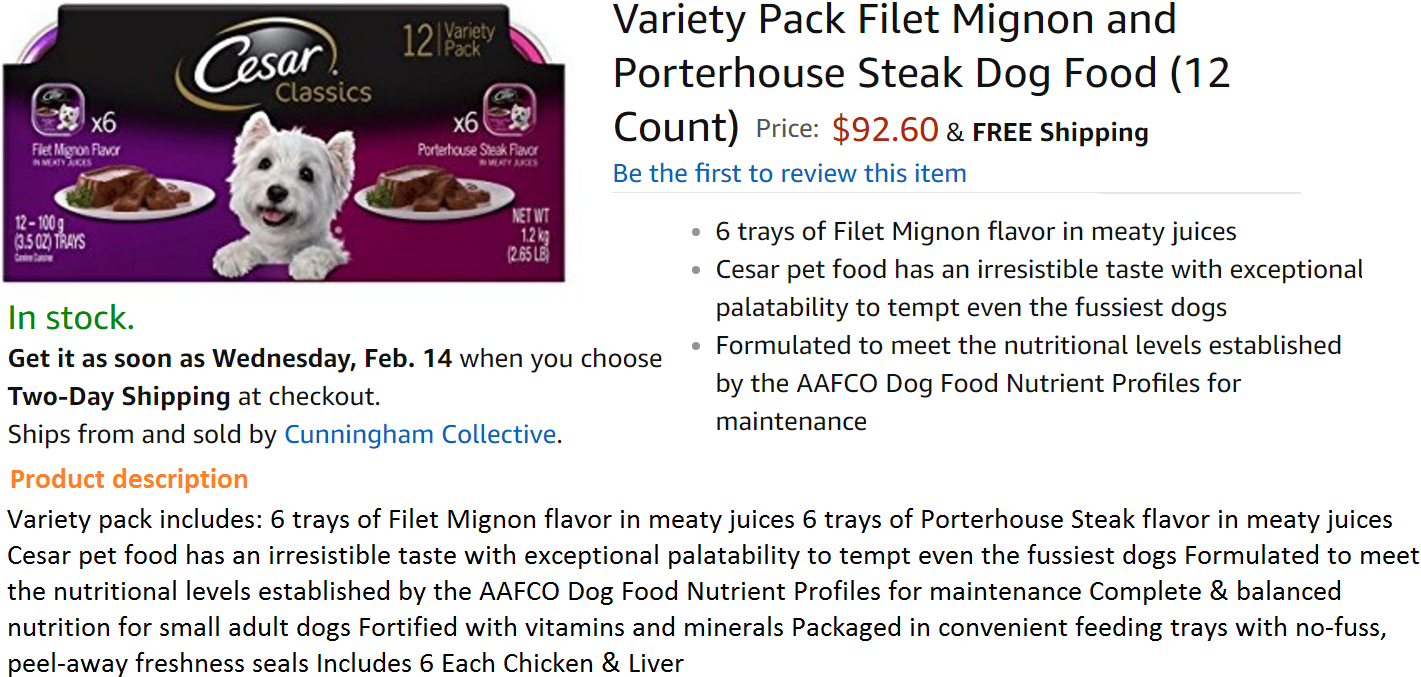}
        \caption{Snapshot of a product profile.}
		\label{fig:profile} 
	\end{center}
\end{figure}

\begin{informal}
  Given a set of target attributes (e.g., brand, flavor, smell), and
  unstructured product profile information like titles, descriptions,
  and bullets: how can we extract values for the attributes from text?
  What if some of these values are new, like emerging brands?
\end{informal}

For a concrete example, refer to Figure~\ref{fig:profile} showing a
snapshot of the product profile of a `dog food' in Amazon.com with
unstructured data such as title, description, and bullets. The product title ``Variety Pack
Fillet Mignon and Porterhouse Steak Dog Food (12 Count)'' contains
{\em two} attributes of interest namely {\em size} and {\em
  flavor}. We want to discover corresponding values for the
attributes like ``12 count" (size), ``Fillet Mignon'' (flavor) and
``Porterhouse Steak'' (flavor).\medskip

\noindent  {\bf Challenges.} This problem presents the following
challenges.

{\em Open World Assumption (OWA).} Previous works for attribute value
extraction~\cite{Ling2012,Putthividhya2011,Petrovski2017,Ghani2006}
work with a {\em closed world assumption} which uses a limited and
pre-defined vocabulary of attribute values. Therefore, these cannot
discover emerging attribute values (e.g., new brands) of newly
launched products that have not been encountered before. OWA renders
traditional multi-class classification techniques an unsuitable choice
to model this problem.

{\em Stacking of attributes and irregular structure.} Product profile information in title and
description is unstructured with tightly packed details about the
product. Typically, the sellers stack several product attributes
together in the title to highlight all important aspects of a
product. Therefore, it is difficult to identify and segment particular
attribute values --- that are often multi-word phrases like ``Fillet
Mignon'' and ``Porterhouse Steak''. Lack of regular grammatical structure renders NLP tools like
parsers, part-of-speech (POS) taggers, and rule-based annotators~\cite{Chiticariu2010, Mikheev1999} less useful. Additionally,
they also have a very sparse context. For instance, over $75$\% of
product titles in our dataset contain fewer than $15$ words whereas over
$60$\% bullets in descriptions contain fewer than 10 words.

{\em Limited Annotated Data.} State-of-the art performance in
attribute value extraction has been achieved by neural networks~\cite{Kozareva2016,Huang2015,Lample2016,ma2016end} which are data
hungry requiring several thousand annotated instances. This does not
scale up with hundreds of product attributes for {\em every} domain,
each assuming several thousand different values. This gives rise to
our second problem statement.
\begin{informal}
  Can we develop supervised models that require limited human
  annotation? Additionally, can we develop models that give
  intepretable explanation for its decisions, unlike black-box methods
  that are difficult to debug?
\end{informal}


\noindent{\bf Contributions.} In this paper, we propose several novel
techniques to address the above challenges. We formulate our problem
as a sequence tagging task similar to named entity recognition
(NER)~\cite{chiu2015named} which has been traditionally used to
identify attributes like names of persons, organizations, and
locations from unstructured text.

We leverage recurrent neural networks like Long Short Term Memory
Networks (LSTM)~\cite{hochreiter1997long} to capture the {\em
  semantics and context} of attributes through distributed word
representations.  LSTM's are a natural fit to this problem because of 
their ability to handle sparse context and sequential nature of the
data where different attributes and values can have
inter-dependencies. 
Although LSTM models capture sequential nature of tokens, they 
overlook the sequential nature of tags. Therefore, we use another sequential model like
conditional random fields (CRF)~\cite{Lafferty2001} to enforce tagging
consistency and extract {\em cohesive} chunks of attribute values
(e.g., multi-word phrases like `fillet mignon') . Although state-of-the-art NER systems~\cite{Kozareva2016,Huang2015,Lample2016,ma2016end} exploit LSTM and CRF, they essentially use them as black box techniques with no explanation. In order to address
the {\em interpretability} challenge, we develop a novel {\em
  attention} mechanism to explain the model's decisions by 
highlighting importance of key concepts relative to their neighborhood
context. Unlike prior works~\cite{Kozareva2016,Huang2015}, OpenTag does not use any dictionary or hand-crafted features.

Neural network models come with an additional challenge that 
they require much more annotated training data than traditional
machine learning techniques because of their huge parameter space. 
Annotation is an expensive task. Therefore, we explore active learning
to reduce the burden of human annotation. 
Overall, we make the following novel contributions.

\begin{itemize} [leftmargin=*]
\item {\bf Model:} We model attribute value extraction as a sequence tagging task
  that supports the Open World Assumption (OWA) and works with
  unstructured text and sparse contexts as in product profiles. We develop a novel model OpenTag leveraging CRF, LSTM, and an {\em
    attention} mechanism to explain its predictions.
\item {\bf Learning:} We explore active learning and novel sampling strategies to
  reduce the burden of human annotation.
\item {\bf Experiments:} We perform extensive experiments in real-life datasets in
  different domains to demonstrate OpenTag's efficacy. It discovers new attribute
  values from as few as $150$ annotated samples (reduction in $3.3$x
  amount of annotation effort) with a high F-score of $83\%$,
  outperforming state-of-the-art models.
\end{itemize}

To the best of our knowledge, this is the first end-to-end framework
for open attribute value extraction addressing key real-world
challenges for modeling, inference, and learning. OpenTag does not make any assumptions about the structure of input text and could be applied to any kind of textual data like profile pages of a  given product.


The rest of the paper is organized as follows: Section 2 presents a
formal description and overview where we introduce
sequence tagging for open attribute value extraction.
Section 3 presents a detailed description of OpenTag using LSTM,
CRF, and a novel attention mechanism. We
discuss active learning strategies in Section 4 followed by extensive
evaluations in real-life datasets in Section 5.  Lastly, Section 6
presents related work followed by conclusions.

%% file: method.tex
\section{Overview}\label{sec:method}

\subsection{Problem Definition}
Given a set of product profiles presented as unstructured text data
(containing information like titles, descriptions, and bullets), and a
set of pre-defined target attributes (e.g., {\em brand}, {\em flavor}, {\em
  size}), our objective is to extract corresponding attribute values
from unstructured text. We have an OWA assumption where we want to
discover new attribute values that may not have been encountered
before. Note that we assume the target attributes (and not attribute-values) for {\em each domain} are given as input to the system. OpenTag automatically figures out the set of applicable attribute-values for each product in the domain.
For example, given the inputs,

\begin{itemize}
\item  target attributes: {\em brand}, {\em flavor}, and {\em size}
\item product title: ``PACK OF 5 - CESAR Canine Cuisine Variety Pack
  Fillet Mignon and Porterhouse Steak Dog Food (12 Count)"
\item product description: ``Variety pack includes: 6 trays of Fillet
  mignon flavor in meaty juices ..."
\end{itemize}

\noindent we want to extract `Cesar' ({\em brand}), `Fillet Mignon'
and `Porterhouse Steak' ({\em flavor}) , and `6 trays' ({\em size}) as
the corresponding values as output from our model. Formally,

\begin{formal}
  Given a set of products $I$, corresponding profiles
  $X=\{x_i: i \in I\}$, and a set of attributes
  $A=\{a_1, \ldots, a_m\}$, extract all attribute-values
  $V_i = \langle \{v_{i, j, 1}, \ldots, v_{i, j, \ell_{i,j}}\}, a_j
  \rangle$
  for $i \in I$ and $j\in [1, m]$ with an open world assumption (OWA);
  we use $v_{i, j}$ to denote {\em the set of values} (of size
  $\ell_{i, j}$) for attribute $a_j$ for the $i^{th}$ product, and the
  product profile (title, description, bullets) consists of a sequence
  of words/tokens $x_i =\{w_{i,1}, w_{i,2}, \cdots w_{i,n_i} \}$.
\end{formal}

Note that we want to discover {\em multiple} values for a given set of 
attributes. For instance, in Figure~\ref{fig:profile} the target
attribute is {\em flavor} and it assumes {\em two} values `fillet
mignon' and `porterhouse steak' for the given product `cesar canine
cuisine'.

\subsection{Sequence Tagging Approach}
A natural approach to cast this problem into a multi-class 
classification problem~\cite{Ling2012} --- treating any target attribute-value as a class label --- 
suffers from the following problems. (1) {\em Label scaling problem:} this method
does not scale well with thousands of potential values for any given
attribute and will increase the volume of annotated training data; (2) {\em Closed world assumption:} it cannot discover any
new value outside the set of labels in the training data; (3) {\em
  Label independence assumption:} it treats each attribute-value
independent of the other, thereby, ignoring any dependency between
them. This is problematic as many attribute-values frequently
co-occur, and the presence of one of them may indicate the presence of
the other. For example, the {\em flavor}-attribute values `fillet
mignon' and `porterhouse steak' often co-occur. Also, the {\em
  brand}-attribute value `cesar' often appears together with the above
{\em flavor}-attribute values.

Based on these observations, we propose a
different approach that models this problem as a {\em sequence tagging
  task}.
\subsubsection{Sequence Tagging}
In order to model the above dependencies between attributes and
values, we adopt the sequence tagging approach. In particular, we
associate a {\em tag} from a given tag-set to {\em each token} in the
input sequence. The objective is to {\em jointly} predict all the tags
in the input sequence. 
In case
of named entity recognition (NER), the objective is to tag {\em
  entities} like (names of) persons, locations, and organizations in
the given input sequence. Our problem is a specific case of NER where
we want to tag attribute values given an input sequence of tokens. The
idea is to exploit {\em distributional semantics}, where similar
sequences of tags for tokens identify similar concepts.

\subsubsection{Sequence Tagging Strategies}
\label{subsec:strategies}
There are several different tagging strategies, where ``BIOE" is the
most popular one.  In BIOE tagging strategy, `B' represents the
beginning of an attribute, `I' represents the inside of an attribute,
`O' represents the outside of an attribute, and `E' represents the end
of an attribute.

Other popular tagging strategies include ``UBIOE" and ``IOB".
``UBIOE" has an extra tag `U' representing the unit token tag that
separates one-word attributes from multi-word ones.  
While for ``IOB"
tagging, `E' is omitted since `B' and `I' are sufficient to express
the boundary of an attribute.

\begin{table}[h!]
\vspace*{-1mm}	
\caption{Tagging Strategies}
\vspace*{-4mm}
	\scriptsize
	\label{tab:tag}
	\begin{center}
		\begin{tabular}{|c|ccccccccc|}
			\hline
			{\bf Sequence}& {\bf duck} & {\bf ,} & {\bf fillet} & {\bf mignon} & {\bf and} & {\bf ranch} & {\bf raised} & {\bf lamb} & {\bf flavor}\\
			\hline
			BIOE& B & O & B &E& O& B& I& E& O\\
			UBIOE& U &O &B &E &O &B &I &E &O\\
			IOB& B &O &B &I &O &B &I &I &O\\
			\hline
		\end{tabular}\vspace*{-2mm}
	\end{center}
\end{table}

Table~\ref{tab:tag} shows an example of the above tagging strategies. 
Given a sequence {\em ``duck, fillet mignon and ranch raised lamb
  flavor"} comprising of 9 words/tokens (including the comma), the
BIOE tagging strategy extracts three {\em flavor}-attributes ``duck",
``fillet mignon" and ``ranch raised lamb" represented by `B', `BE' and
``BIE" respectively. 

\subsubsection{Advantages of Sequence Tagging} The sequence tagging
approach enjoys the following benefits. (1) {\em OWA and label scaling.} A tag is associated to a token, and not a
specific attribute-value, and, therefore scales well with new values.
(2) {\em Discovering multi-word attribute values.} The above strategy
extracts {\em sequence} of tokens (i.e., multi-word values) as opposed to
identifying single-word values. (3) {\em Discovering multiple attribute values.} The tagging strategy can
be extended to discover values of multiple attributes at the same time
if they are tagged differently from each other. For instance,
to discover two attributes `flavor' and `brand' {\em jointly}, we can
tag the given sequence with tags as {\small `Flavor-B', `Flavor-I'
,`Flavor-O', and `Flavor-E'} to distinguish from {\small `Brand-B', `Brand-I',
`Brand-O', and `Brand-E'}.

\noindent{\bf Formulation of our approach.}
Following the above discussions, we can reduce our original problem of
{\sf\em Open Attribute Value Extraction} to the following {\sf\em Sequence
  Tagging Task}.

Let ${Y}$ be the tag set containing all the tags decided by the
tagging strategy.  If we choose BIOE as our tagging strategy, then
${Y}=\left\{B, I, O, E\right\}$.  Tag set of other strategies can be
derived following a similar logic. Our objective is to learn a tagging
model $F\left(x\right){\rightarrow}y$ that assigns each token
$w_{ij} \in W$ of the input sequence $x_i \in X$ of the $i^{th}$
product profile with a corresponding tag $y_{ij} \in Y$.  The training
set for this supervised classification task is given by
$\mathcal{S}=\left\{\left(\mathbf{x}_i,\mathbf{y}_i\right)\right\}_{i=1}^T$.
This is a global tagging model that captures relations between tags
and models the entire sequences as a whole. We denote our framework by
\sysname.

%% file: base.tex
\section{\sysname Model: Extraction via Sequence Tagging}\label{sec:base}
OpenTag builds upon state-of-the-art named entity recognition (NER) systems~\cite{Kozareva2016,Huang2015,Lample2016,ma2016end} that use bidirectional LSTM and conditional random fields, but without using any dictionary or hand-crafted features as in~\cite{Kozareva2016,Huang2015}. In the following section, we will first review these building blocks, and how we {\em adapt} them for attribute value extraction.
Thereafter, we outline our novel contributions of using {\em Attention}, end-to-end OpenTag architecture, and active learning to reduce requirement of annotated data.


\subsection{Bidirectional LSTM (BiLSTM) Model}
Recurrent neural networks (RNN) capture long range dependencies between tokens in a
sequence. Long Short Term Memory Networks (LSTM) were developed to address the vanishing gradient problems of RNN. A basic
LSTM cell consists of various gates to control the flow of information through
the LSTM connections. By construction, LSTM's are suitable for sequence tagging
or classification tasks where it is insensitive to the gap length
between tags unlike RNN or Hidden Markov Models (HMM). 

Given an input $e_t$ (say, the word embedding of token $x_t \in X$), an LSTM cell performs various non-linear transformations to generate a hidden vector state $h_t$ for each token at each timestep $t$ that can be later used for tasks such as classification.

Bidirectional LSTM's are an improvement over LSTM that capture both
the previous timesteps (past features) and the future timesteps
(future features) via forward and backward states respectively. In
sequence tagging tasks, we often need to consider both the left and
right contexts {\em jointly} for a better prediction
model. Correspondingly, two LSTM's are used, one with the standard
sequence, and the other with the sequence reversed. Correspondingly,
there are two hidden states that capture past and future information
that are concatenated to form the final output.

Using the hidden vector representations from {\em forward} and {\em
  backward} LSTM ($\overrightarrow{h_t}$ and $\overleftarrow{h_t}$
respectively) along with a non-linear transformation, we can create a
new hidden vector as:
$h_t = \sigma ([\overrightarrow{h_t}, \overleftarrow{h_t}])$.


Finally, we add a softmax function to predict the tag for each token
$x_t$ in the input sequence $x = \langle x_t \rangle$ given hidden
vector $\langle h_t \rangle$ at each timestep:
\begin{equation}
\Pr(y_t=k)=\softmax(h_t \cdot W_h), 
\end{equation}

\noindent where $W_h$ is the variable matrix shared across all 
tokens, and $k \in \{B,I,O,E\}$. For each token, the tag with the
highest probability is generated as the output tag. Using the
ground-labels we can train the above BiLSTM network to learn all 
parameters $W$ and $H$ using backpropagation. 

{\noindent \bf Drawbacks for sequence tagging:} The BiLSTM model considers {\em sequential nature of
	the given input sequence, but not the output tags}. As a result, the above model does not consider the coherency of tags
during prediction. The prediction for each tag is made independent of
the other tags.  For example,
given our set of tags $\{B, I, O, E\}$, the model may predict a
mis-aligned tag sequence like $\{B, O, I, E\}$, leading to an
incoherent attribute extraction. In order to avert this problem, we
use Conditional Random Fields (CRF) to {\em also consider the
  sequential nature of the predicted tags}.

\subsection{Tag Sequence Modeling with Conditional Random Fields and
  BiLSTM}
\subsubsection{Conditional Random Fields (CRF)}
For sequence labeling tasks, it is important to consider the association
or correlation between labels in a neighborhood, and use this
information to predict the best possible label sequence given an input
sequence. For example, 
if we already know the starting
boundary of an attribute (B), this increases the likelihood of the
next token to be an intermediate (I) one or end of boundary (E),
rather than being outside the scope of the attribute (O). {\em
  Conditional Random Fields (CRF) allows us to model the label sequence
  jointly}.


Given an input sequence $x=\{x_1, x_2, \cdots x_n \}$ and
corresponding label sequence $y = \{y_1, y_2, \cdots y_n\}$, the joint
probability distribution function for the CRF can be written as the
conditional probability:
\begin{equation*}
  \Pr(y|x; \Psi) \propto exp \bigg( \sum_{k=1}^K \psi_k f_k (y, x) \bigg),
\end{equation*}

\noindent where $f_k(y, x)$ is the feature function, $\psi_K$ is the
corresponding weight to be learned, $K$ is the number of features, and
$\mathcal{Y}$ is the set of all possible labels. Traditional NER
leverages several user defined features based on the current and
previous token like the presence of determiner (`the'), presence of
upper-case letter, POS tag of the current token (e.g.,
`noun') and the previous  (e.g., `adjective'), etc.

Inference for general CRF is intractable with a complexity of $|\mathcal{Y}|^n$ where $n$ is the
length of the input sequence and $|\mathcal{Y}|$ is the cardinality of
the label set. We use linear-chain
CRF's to avoid this problem. We constrain the feature functions to
depend only on the neighboring tags $y_t$ and $y_{t-1}$ at timestep
$t$. This reduces the computational complexity to
$|\mathcal{Y}|^2$. We can re-write the above equation as: 
\begin{equation*}
\Pr(y|x; \Psi) \propto \prod_{t=1}^T exp \bigg( \sum_{k=1}^K \psi_k f_k (y_{t-1}, y_t, x) \bigg).
\end{equation*}

\subsubsection{Bidirectional LSTM and CRF Model}
As we described above, traditional CRF models use several manually
defined syntactic features for NER tasks. In this work, we combine
LSTM and CRF to use semantic features like the distributed word
representations. We do not use any hand-crafted features like in prior works~\cite{Kozareva2016,Huang2015}. Instead, the hidden states generated by the BiLSTM model are used as input features for the CRF model. We incorporate an
additional non-linear layer to weigh the hidden states that capture
the importance of different states for the final tagging decision.

The BiLSTM-CRF network can use (i) features from the previous as well
as future timesteps, (ii) semantic information of the given input
sequence encoded in the hidden states via the BiLSTM model, and (iii)
tagging consistency enforced by the CRF that captures dependency
between the output tags. The objective now is to predict the best
possible tag sequence of the entire input sequence given the hidden
state information $\langle h_t \rangle$ as features to the CRF. The
BiLSTM-CRF network forms the second component for our model.
\begin{equation*}
\Pr(y|x; \Psi) \propto \prod_{t=1}^T exp \bigg( \sum_{k=1}^K \psi_k f_k (y_{t-1}, y_t, \langle h_t \rangle) \bigg).
\end{equation*}

\subsection{\sysname: Attention Mechanism}
In this section, we describe our novel attention mechanism that can be used to explain the model's tagging decision unlike the prior NER systems~\cite{Kozareva2016,Huang2015,Lample2016,ma2016end} that use BiLSTM-CRF as black-box.

In the above BiLSTM-CRF model, we consider all the hidden states
generated by the BiLSTM model to be important when they are used as
features for the CRF. However, not all of these states are equally
important, and some mechanism to make the CRF aware of the important
ones may result in a better prediction model. This is where {\em
  attention} comes into play.

The objective of the attention mechanism is to highlight important
concepts, rather than focusing on all the information. Using such
mechanism, we can highlight the important tokens in a given input
sequence responsible for the model's predictions as well as performing
feature selection. This has been widely used in the vision community
to focus on a certain region of an image with ``high resolution''
while perceiving the surrounding image in ``low resolution'' and then
adjusting the focal point over time.

In the Natural Language Processing domain, attention mechanism has
been used with great success in Neural Machine Translation
(NMT)~\cite{Bahdanau2014}. NMT systems comprise of a
sequence-to-sequence encoder and decoder.  Semantics of a sentence is
mapped into a fixed-length vector representation by an encoder, and
then the translation is generated based on that vector by a decoder.
In the original NMT model, the decoder generates a translation solely
based on the last hidden state.  But it is somewhat unreasonable to
assume all information about a potentially very long sentence can be
encoded into a single vector, and that the decoder will produce a good
translation solely based on that.  With an attention mechanism, instead
of encoding the full source sequence into a fixed-length vector, we
allow the decoder to {\em attend} to different parts of the source
sentence at each step of the output generation.  Importantly, we let
the model learn what to attend to based on the input sentence and what
it has produced so far.

We follow a similar idea. In our setting, the encoder is the
underlying BiLSTM model generating the hidden state representation
$\langle h_{t} \rangle$.  We introduce an attention layer with an
attention matrix $A$ to capture the similarity of any token with
respect to all the neighboring tokens in an input sequence. The
element $\alpha_{t,t'} \in A$ captures the similarity between the
hidden state representations $h_{t}$ and $h_{t'}$ of tokens $x_{t}$
and $x_{t'}$ at timesteps $t$ and $t'$ respectively. The attention
mechanism is implemented similar to an LSTM cell as follows:

\begin{align}
g_{t,t'} &= tanh(W_g h_{t} + W_{g'} h_{t'} + b_g),\\
\alpha_{t,t'} &= \sigma(W_a g_{t,t'} + b_a),
\end{align}

\noindent where, $\sigma$ is the element-wise sigmoid function, $W_g$ and
$W_{g'}$ are the weight matrices corresponding to the hidden states
$h_{t}$ and $h_{t'}$; $W_a$ is the weight matrix corresponding to
their non-linear combination; $b_g$ and $b_a$ are the bias vectors.

The attention-focused hidden state representation $l_t$ of a token at
timestep $t$ is given by the weighted summation of the hidden state
representation $h_{t'}$ of all other tokens at timesteps $t'$, and
their similarity $\alpha_{t,t'}$ to the hidden state representation
$h_t$ of the current token. Essentially, $l_t$ dictates how much to
{\em attend} to a token at any timestep {\em conditioned on their
  neighborhood context}. This can be used to highlight the model's final tagging
decision based on token importance.  
\begin{equation}
 l_t = \sum_{t'=1}^n \alpha_{t,t'} \cdot h_{t'}.
\end{equation}

\noindent In Section~\ref{subsec:attention-expt}, we discuss how OpenTag generates interpretable explanations of its tagging decision using this attention matrix.

\subsection{Word Embeddings}
\label{subsec:word-embed}
Neural word embeddings map words that co-occur in a similar context to
nearby points in the embedding space~\cite{Mikolov2013}. This forms the first layer of our architecture. Compared to
bag-of-words (BOW) features, word embeddings capture both syntactic
and semantic information with low-dimensional and dense word
representations. The most popular tools for this purpose are
Word2Vec~\cite{Mikolov2013} and GloVe~\cite{pennington2014glove}, which
are trained over large unlabeled corpus. Pre-trained embeddings have a single representation
for each token. This does not serve our purpose as the same word
can have a different representation in different contexts. For
instance, `duck' (bird) as a {\em flavor}-attribute value should have
a different representation than `duck' as a {\em brand}-attribute
value. Therefore, we learn the word representations 
conditioned on the attribute tag (e.g., `flavor'), and generate
different representations for different attributes. In our setting,
each token at time $t$ is associated with a vector $e_t \in R^d$, where
$d$ is the embedding dimension. The elements in the vector are latent,
and considered as parameters to be learned.

\subsection{\sysname Architecture: Putting All Together}

\begin{figure}[htbp]
	\begin{center}
		\includegraphics[width=1.0\linewidth]{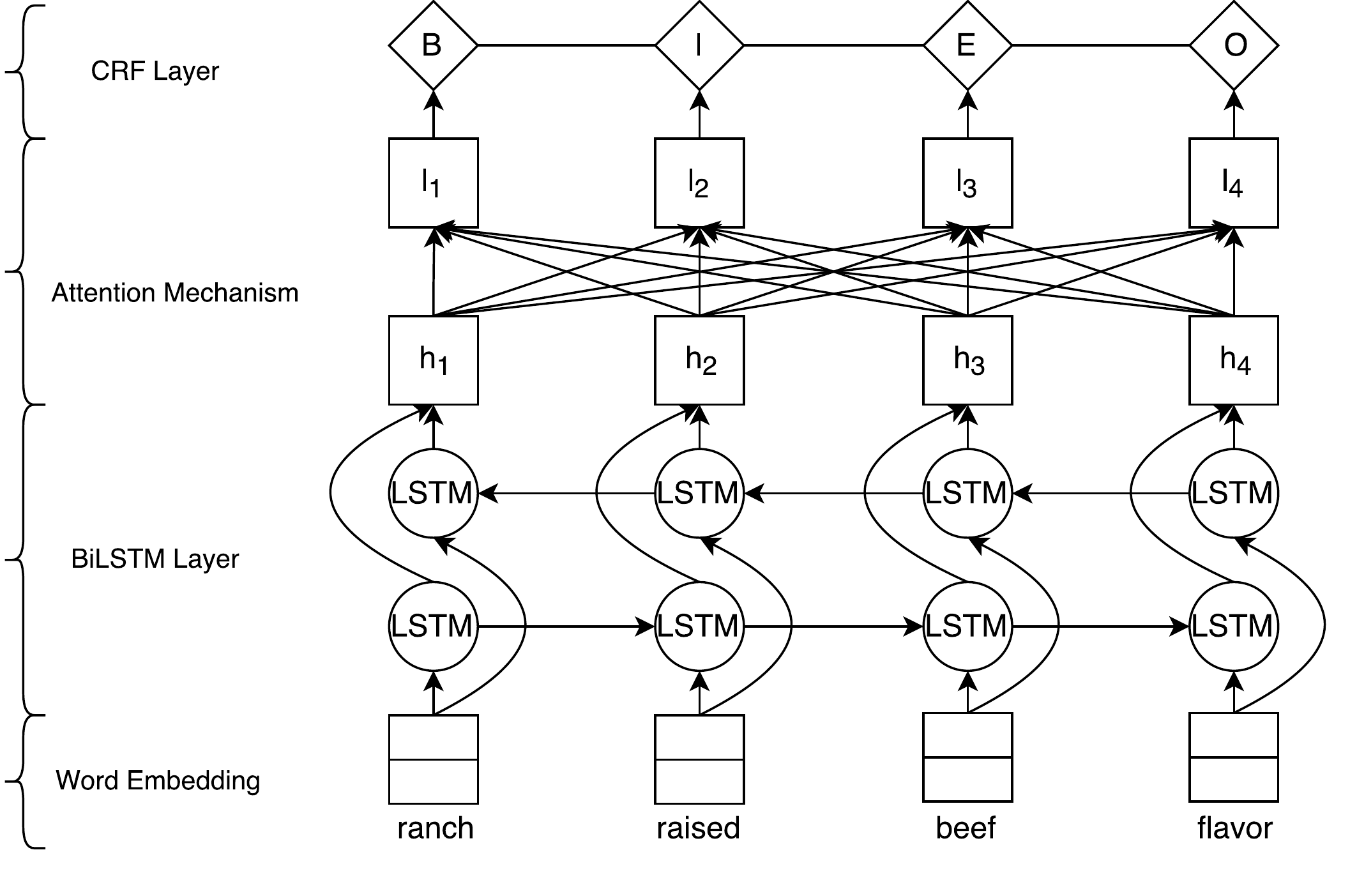}
\vspace*{-6mm}		
\caption{\small \sysname Architecture: BiLSTM-CRF with Attention.
}
		\label{fig:arch} \vspace*{-4mm}
	\end{center}
\end{figure}

Figure~\ref{fig:arch} shows the overall architecture of \sysname. The
first layer is the word embedding layer that generates an embedding
vector $e_t$ for each token $x_t$ in the input sequence $x$. This
vector is used as an input to the bidirectional LSTM layer
that generates its hidden state representation $h_t$ as a
concatenation of the forward and backward LSTM states. This representation captures
its future and previous timestep features.

The output of BiLSTM goes as input to the attention layer that learns which states to
focus or {\em attend} to in particular --- generating the
attention-focused hidden state representation $\langle l_t \rangle$
for the input sequence $\langle x_t \rangle$. These representations are used as input
features in the CRF that enforces tagging consistency --- considering
dependency between output tags and the hidden state representation of
tokens at each timestep. The joint probability distribution of the tag
sequence is given by:
\begin{equation}\label{eq:6}
  \Pr(y|x; \Psi) \propto \prod_{t=1}^T exp \bigg( \sum_{k=1}^K \psi_k f_k (y_{t-1}, y_t, \langle l_t \rangle) \bigg).
\end{equation}

For training this network, we use the maximum conditional likelihood
estimation, where we maximize the log-likelihood of the above joint
distribution with respect to all the parameters $\Psi$ over $m$
training instances
$\left\{\left(\mathbf{x}_i,\mathbf{y}_i\right)\right\}_{i=1}^m$:
\begin{equation}
	L(\Psi) = \sum_{i=1}^m \log\Pr(\mathbf{y}_i|\mathbf{x}_i; \Psi).
\end{equation}

The final output is the best possible tag sequence $y^{*}$ with the
highest conditional probability given by:
\begin{equation}
	y^{*} = \argmax_y\Pr(y|x; \Psi).
\end{equation}

%% file: active.tex
\section{\sysname: Active Learning}\label{sec:active}
In this section, we present our novel active learning framework for OpenTag to reduce the burden of human annotation. 

An essential requirement of supervised machine learning algorithms is
annotated data. However, manual annotation is expensive and time
consuming. In many scenarios, we have access to a lot of unlabeled
data. Active learning is useful in these scenarios, where we can allow
the learner to select samples from the un-labeled pool of data, and
request for labeling.

Starting with a {\em small set of labeled instances} as an initial
training set $L$, the learner iteratively requests labels for one or
more instances from a {\em large unlabeled pool} of instances $U$
using some query strategy $Q$. These instances are labeled, and added
to the base set $L$, and the process is repeated till some stopping
criterion is reached. The challenge is to design a good query strategy
$Q$ that selects the most informative samples from $U$ given the
learner's hypothesis space. This aims to improve the learner's
performance with {\em as little annotation effort as possible}. This
is particularly useful for sequence labeling tasks, where the
annotation effort is proportional to the length of a
sequence, in contrast to instance classification tasks. \sysname
employs active learning with a similar objective to reduce manual
annotation efforts, while making judicious use of the large number of
unlabeled product profiles. There are many different approaches to
formulate a query strategy to select the most informative instances to
improve the active learner. 

As our baseline strategy, we consider the method of least confidence (LC)~\cite{Culotta2005} which is shown to perform quite well in practice~\cite{Settles2008}. It selects the sample for which the classifier 
is least confident. 
In our sequence tagging task,
the confidence of the CRF in tagging an input sequence is given by the
conditional probability in Equation~\ref{eq:6}. Therefore, the query
strategy selects the sample $x$ with maximum uncertainty given by:
\begin{equation}
	Q^{lc}(x) = 1 - \Pr(y^{*}|x; \Psi),
\end{equation}
where $y^{*}$ is the best possible tag sequence for $x$. 

However, this strategy has the following drawbacks: (1) The conditional probability of the entire sequence is proportional to
the product of (potential of) successive tag
($\langle y_{t-1}, y_t \rangle$) transition scores. Therefore, a false
certainty about any token's tag $y_t$ can pull down the probability of
the entire sequence --- leading to missing a valuable query. 
(2) When the oracle reveals the tag of a token, this may impact only a few other tags, having a relatively low impact on the entire sequence. 

\subsection{Method of Tag Flips}
In order to address these limitations, we formulate a new query
strategy to identify informative sequences {\em based on how difficult
  it is to assign tags to various tokens in a sequence}.

In this setting, we simulate a committee of \sysname learners
$C= \{\Psi^{(1)}, \Psi^{(2)}, \cdots \Psi^{(E)}\}$ to represent
different hypotheses that are consistent with labeled set $L$. The most informative sample is the one for which there is major
disagreement among committee members. 

We train \sysname for a preset number of epochs $E$ using dropout \cite{srivastava2014} regularization technique. Dropout prevents overfiting during  network training by randomly dropping units in the network with their connections. Therefore, for each epoch
$e$, \sysname learns a different set of models and parameters $\Psi^{(e)}$ ---
thereby simulating a committee of learners due to the dropout mechanism.

After each epoch, we apply
$\Psi^{(e)}$ to the unlabeled pool of samples and record the best
possible tag sequence $y^{*}(\Psi^{(e)})$ assigned by the learner to
each sample.

We define a {\em flip} to be a change in the tag of a token of a given
sequence across {\em successive} epochs, i.e., the learners
$\Psi^{(e-1)}$ and $\Psi^{(e)}$ assign different tags
$y^{*}_t(\Psi^{(e-1)}) \ne y^{*}_t(\Psi^{(e)})$ to token $x_t \in x$.
If the tokens of a given sample sequence frequently change tags across successive epochs, this indicates \sysname is uncertain about the
sample, and {\em not stable}. Therefore, we consider tag flips (TF) to be a measure of
uncertainty for a sample and model stability, and {\em query for labels for the samples
  with the highest number of tag flips}.\smallskip

\begin{table}
	\vspace{-2mm}
	\caption{{\small Sampling Strategies. LC: Least confidence. TF: Tag flip.}}
	\label{tab:sampling}\vspace{-4mm}
	\scriptsize
	\begin{center}
		\begin{tabular}{|c|ccccccccc|}
			\hline
			& duck &, & fillet & mignon & and & ranch & raised & lamb & flavor\\
			\hline
			Gold-Label (G) & B & O & B &E& O& B& I& E& O\\
			Strategy: LC (S1) & O & O & B &E& O& B& I& E& O\\
			Strategy: TF (S2) & B & O & B &O& O& O & O& B& O\\
			\hline
		\end{tabular}\vspace{-4mm}
	\end{center}
\end{table}

\noindent {\bf Illustrative example.} Consider the snippet in Table~\ref{tab:sampling} and the tag sequences assigned by two different sampling strategies $S1$ and $S2$ corresponding to least confidence and tag flip respectively.  Assume that the gold-label sequence (G) is known. In practice, we do {\em not} use the ground labels for computing tag-flips during learning; instead we use predictions from OpenTag from the previous epoch. 

Contrasting the tag
sequence $S2$ with the gold sequence $G$, we observe $4$ flips
corresponding to mismatch in tags of `mignon' and `ranch raise lamb'.
 
Given an unlabeled pool of instances, the strategy for least
confidence may pick sequence $S1$ that the learner is most uncertain
about in terms of the overall probability of the entire sequence. We
observe this may be due to mis-classifying `duck' that is an important
concept at the start of the sequence. However, the learner gets the
remaining tags correct. In this case, if the oracle assigns the tag for `duck', it
does not affect any other tags of the sequence. Therefore, this is not
an informative query to be given to the oracle for labeling.

On the other hand, the tag flip strategy selects sequence $S2$ based
on the number of flips of token-tags that the model has grossly mis-tagged. Labeling
this query has much more impact on the learner to tune its
parameters than the other sequence $S1$. 

The flip based sampling
strategy is given by:
\begin{equation}
Q^{tf}(x) = \sum_{e=1}^E \sum_{t=1}^n\mathcal{I}(y^{*}_t(\Psi^{(e-1)}) \ne y^{*}_t(\Psi^{(e)})),
\end{equation}
where $y^{*}_t(\Psi^{(e)})$ is the best possible tag sequence for $x$
assigned by the learner $\Psi^{(e)}$ in epoch $e$ and
$\mathcal{I(\cdot)}$ is an indicator function that assumes the value
$1$ when the argument is true, and $0$ otherwise. $Q^{tf}$ computes the number of tag flips of tokens of a sequence across successive epochs.

Algorithm~\ref{algo:active} outlines our active learning process. The
batch-size indicates how many samples we want to query for
labels. Given a batch-size of $B$, the top $B$ samples with the
highest number of flips are manually annotated with tags. We continue the active learning process until the validation loss converges within a threshold.

{
 \begin{algorithm}[t]
 	\small
	\SetAlgoLined
	\DontPrintSemicolon
		Given: Labeled set $L$, unlabeled pool $U$, query strategy $Q$, query batch size $B$ \;
		\Repeat {some stopping criterion} {
		\For {each epoch $e \in E$} {
			// simulate a committee of learners using current $L$ \;
			$\Psi^{(e)}$ = train($L$) \;
			Apply $\Psi^{(e)}$ to unlabeled pool $U$ and record tag flips
		}	
		\For {each query $b \in B$} {
		// find the instances with most tag flips over $E$ epochs \;
		   $x^{*} = \argmax_{x \in U} Q^{tf}(x)$ \;
		// label query and move from unlabeled pool to labeled set \;
		 $L = L \cup \{x^{*}, label(x^{*})\}$ \;
		 $U = U - x^{*}$\;
		}	
}	
	\caption{Active learning with tag flips as query strategy.}
	\label{algo:active}
\end{algorithm}

%% file: experiment.tex
\section{Experiments}\label{sec:experiment}
\subsection{\sysname: Training}
We implemented \sysname using Tensorflow, where some basic layers are
brought from Keras.\footnote{Note that we do not do any
  hyper-parameter tuning. Most default parameter values come from
  Keras. It may be possible to boost OpenTag performance by careful
  tuning.}. We run our experiments within docker containers on a 72-core machine powered by Ubuntu Linux. 



We use $100$-dimensional pre-trained word vectors from
GloVe~\cite{pennington2014glove} for initializing our word embeddings
that are optimized during training. Embeddings for words not in GloVe are randomly initialized and re-trained. Masking is adopted to support variable length input. We set the hidden size of LSTM to $100$, which
generates a $200$ dimensional output vector for BiLSTM after
concatenation.  The dropout rate is set to $0.4$. We use
Adam~\cite{kingma2014} for parameter optimization with a batch size of
$32$. We train the models for $500$ epochs, and report the averaged
evaluation measures for the last $20$ epochs.


\subsection{Data Sets}
We perform experiments in $3$ domains, namely, (i) dog food, (ii)
detergents, and (iii) camera. For each domain, we use the product
profiles (like titles, descriptions, and bullets) from {\tt
  Amazon.com} public pages. The set of applicable attributes are defined per-domain. For each product in a domain, OpenTag figures out the set of applicable attribute values. We perform experiments with different configurations to
validate the robustness of our model.


Table~\ref{tab:dataset} gives the description of different data sets
and experimental settings. It shows the (i) domain, (ii) type of
profile, (iii) target attribute, (iv) number of samples or products we
consider, and (v) the number of extractions in terms of
attribute values. `Desc' denotes description whereas `Multi' refers to
multiple attributes (e.g., flavor, capacity, and brand). `DS'
represents a disjoint training and test set with no overlapping
attribute values; for all other data sets, we randomly split them into
training and test instances.\todo{insert a note on how do you split
  them i.e. any specific ratio?}

\begin{table}
	\caption{Data sets.}
	\vspace{-4mm}
	\scriptsize
	\label{tab:dataset}
	\begin{center}
		\begin{tabular}{|c|c|c|c|c|c|c|}
			\hline
			Domain & Profile & Attribute & \multicolumn{2}{c|}{Training} & \multicolumn{2}{c|}{Testing}\\
			\cline{4-7}
			& & & Samples & Extractions &  Samples &  Extractions\\
			\hline
			Dog Food (DS)& Title & Flavor & 470 & 876 & 493 & 602 \\
			\hline
			Dog Food & Title & Flavor & 470 & 716 & 493 & 762 \\
			&Desc & Flavor & 450 & 569 & 377 & 354  \\
			&Bullet & Flavor & 800 & 1481 & 627 & 1179 \\
			&Title & Brand & 470 & 480 & 497 & 607 \\
			&Title & Capacity & 470 & 428 & 497 & 433 \\
			&Title & Multi & 470 & 1775 & 497 & 1632 \\
			\hline
			Camera & Title & Brand & 210 & 210 & 211 & 211 \\
			\hline
			Detergent & Title & Scent & 500 & 487 & 500 & 484 \\
			\hline
		\end{tabular}
	\end{center}
\vspace{-1em}
\end{table}

\noindent{\bf Evaluation measure.} We evaluate the {\em precision},
{\em recall}, and {\em f-score} of all models. In contrast to prior
works evaluating tag-level measures --- we evaluate extraction quality
of our model with either full or no credit. In other words, given a target {\em
  flavor}-extraction of ``ranch raised lamb'', a model gets credit
{\em only} when it extracts the full sequence.  After a model assigns
the best possible tag decision, attribute values are extracted and
compared with ground truth.

\subsection{Performance: Attribute Value Extraction}
{\noindent \bf Baselines.} The first baseline we consider is the
BiLSTM model~\cite{hochreiter1997long}. The second one is the
state-of-the-art sequence tagging model for named entity recognition
(NER) tasks using BiLSTM and
CRF~\cite{Kozareva2016,Huang2015,Lample2016,ma2016end} but without
using any dictionary or hand-crafted features as
in~\cite{Kozareva2016,Huang2015} . We adopt these models for attribute
value extraction. Training and test data are the same for all the
models.\smallskip

{\noindent \bf Tagging strategy.} Similar to previous works, we also
adopt $\{B,I,O,E\}$ tagging strategy. We experimented with other
tagging strategies, where $\{B,I,O,E\}$ performed marginally better
than the others.\smallskip

\begin{table}
	\centering
	\small
	\caption{Performance comparison of different models on attribute value extraction 
      for different product profiles and datasets. OpenTag outperforms other state-of-the-art NER systems 
      \cite{Kozareva2016,Huang2015,Lample2016,ma2016end} based on BiLSTM-CRF.} 
\vspace{-1em}
	\begin{tabular}{p{.7cm}p{1.7cm}p{.4cm}p{.4cm}p{.4cm}}
		\toprule
		\multicolumn{1}{l}{Datasets/Attribute} & Models & \multicolumn{1}{l}{Precision} & \multicolumn{1}{l}{Recall} & \multicolumn{1}{l}{Fscore} \\\midrule
		\multicolumn{1}{l}{Dog Food: Title} & BiLSTM & 83.5  & 85.4  & 84.5 \\
		\multicolumn{1}{l}{Attribute: Flavor} & BiLSTM-CRF & 83.8  & 85.0  & 84.4 \\
		& OpenTag & {\bf 86.6}  & {\bf 85.9}  & {\bf 86.3} \\\midrule
		\multicolumn{1}{l}{Camera: Title} & BiLSTM & 94.7  & 88.8  & 91.8 \\
		\multicolumn{1}{l}{Attribute: Brand} & BiLSTM-CRF & 91.9  & {\bf 93.8}  & 92.9 \\
		& OpenTag & {\bf 94.9}  & 93.4  & {\bf 94.1} \\\midrule
		\multicolumn{1}{l}{Detergent: Title} & BiLSTM & 81.3  & 82.2  & 81.7 \\
		\multicolumn{1}{l}{Attribute: Scent} & BiLSTM-CRF & {\bf 85.1}  & 82.6  & 83.8 \\
		& OpenTag & 84.5  & {\bf 88.2}  & {\bf 86.4} \\\midrule
		\multicolumn{1}{l}{Dog Food: Description} & BiLSTM & 57.3  & 58.6  & 58 \\
		\multicolumn{1}{l}{Attribute: Flavor} & BiLSTM-CRF & 62.4  & 51.5  & 56.9 \\
		& OpenTag & {\bf 64.2}  & {\bf 60.2}  & {\bf 62.2} \\\midrule
		\multicolumn{1}{l}{Dog Food: Bullet} & BiLSTM & 93.2  & 94.2  & 93.7 \\
		\multicolumn{1}{l}{Attribute: Flavor} & BiLSTM-CRF & 94.3  & 94.6  & 94.5 \\
		& OpenTag & {\bf 95.7}  & {\bf 95.7}  & {\bf 95.7} \\\midrule
		\multicolumn{1}{l}{Dog Food: Title} & BiLSTM & 71.2  & 67.4  & 69.3 \\
		\multicolumn{1}{l}{Multi Attribute:} & BiLSTM-CRF & 72.9  & 67.3  & 70.1 \\
		\multicolumn{1}{l}{Brand, Flavor, Capacity}  & OpenTag & {\bf 76.0}    & {\bf 68.1}  & {\bf 72.1} \\\bottomrule
	\end{tabular}
	\label{tab:overall-perf}
\end{table}

{\noindent \bf Attribute value extraction results.} We compare the
performance of \sysname with the aforementioned baselines for identifying
attribute values from different product profiles (like title,
description and bullets) and different sets of attributes (like brand,
flavor and capacity) in different domains (like dog food, detergent and
camera). Table~\ref{tab:overall-perf} summarizes the results, where all
experiments were performed on random train-test splits. The first
column in the table shows the domain, profile type, and attribute we
are interested in. We observe that \sysname consistently outperforms
competing methods with a high overall f-score of $82.8\%$.

We also observe the highest performance improvement ($5.3\%$) of
\sysname over state-of-the-art BiLSTM-CRF model on product
descriptions, which are more structured and provide more context
than either titles or bullets. However, overall performance of
\sysname for product descriptions is much worse. Although descriptions
are richer in context, the information present is also quite diverse
in contrast to titles or bullets that are short, crisp and focused.\smallskip

\begin{table}
	\centering
	\small
	\caption{OpenTag results on disjoint split; where it discovers new attribute values never seen before with $82.4\%$ f-score.}
				\vspace{-1em}
	\begin{tabular}{lrrr}
		\toprule
		Train-Test Framework & \multicolumn{1}{l}{Precision} & \multicolumn{1}{l}{Recall} & \multicolumn{1}{l}{F-score} \\\midrule
		Disjoint Split (DS) & 83.6  & 81.2  & 82.4 \\
		Random Split & 86.6  & 85.9  & 86.3 \\\bottomrule
	\end{tabular}
	\label{tab:owa}
\end{table}

{\noindent \bf Discovering new attribute values with open world
  assumption (OWA).} In this experiment (see Table \ref{tab:owa}), we
want to find the performance of \sysname in discovering new attribute
values it has {\em never seen} before. Therefore, we make a clear
separation between training and test data such that they do not share
any attribute value. Compared with the earlier random split setting, \sysname still performs well in the disjoint
setting with a f-score of $82.4\%$ in discovering new attribute
values for {\em flavors} from dog food titles. However, it is worse than random split -- where it has the
chance to see some attribute values during training, leading to
better learning.\smallskip



\begin{table}
	\centering
	\small
	\caption{OpenTag has improved performance on extracting values of multiple attributes jointly vs. single extraction.}
				\vspace{-1em}
	\begin{tabular}{lrrr}
		\toprule
		Attribute  & \multicolumn{1}{l}{Precision} & \multicolumn{1}{l}{Recall} & \multicolumn{1}{l}{F-Score} \\\midrule
		Brand: Single & 52.6  & 42.6  & 47.1 \\
		Brand: Multi & {\bf 58.4}  & {\bf 44.7}  & {\bf 50.6} \\\midrule
		Flavor: Single & 83.6  & {\bf 81.2}  & {\bf 82.4} \\
		Flavor: Multi & {\bf 83.7}  & 77.5  & 80.5 \\\midrule
		Capacity: Single & 81.5  & 86.4  & 83.9 \\
		Capacity: Multi & {\bf 87.0}    & {\bf 87.2}  & {\bf 87.1} \\\bottomrule
	\end{tabular}
	\label{tab:multi-attr}
	\vspace{-1em}
\end{table}

{\noindent \bf Joint extraction of multi-attribute values.} As we
discussed in Section~\ref{subsec:strategies}, \sysname is able to
extract values of multiple attributes jointly by modifying the tagging
strategy. In this experiment, we extract values for {\em brand,
  flavor} and {\em capacity} from titles of dogfood data {\em jointly} on a disjoint split of data. Using $\{B,I,O,E\}$ as the
tagging strategy, each attribute $a$ has its own $\{B_a, I_a, E_a\}$
tag with $O$ shared among them -- with a total of $10$ tags for three
attributes. From Table~\ref{tab:overall-perf}, we observe \sysname to
have a $2\%$ f-score improvement over our strongest BiLSTM-CRF
baseline.

As we previously argued, {\em joint} extraction of multiple attributes
can help leverage their distributional semantics together, thereby,
improving the extraction of {\em individual} ones as shown in
Table~\ref{tab:multi-attr}. \todo{we do not do well for flavor in this
  case. if there is any insight insert here} Although the performance
in extracting {\em brand} and {\em capacity} values improve in the
joint setting, the one for {\em flavor} marginally degrades.

\subsection{\sysname: Interpretability via Attention}
\label{subsec:attention-expt}

\begin{figure}[t]
	\begin{center}
		\vspace{-3.5em}
		\includegraphics[width=0.7\linewidth]{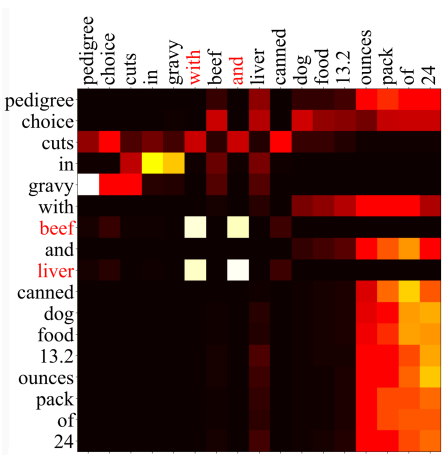}
		\vspace{-4em}
		\caption{\small \sysname shows interpretable explanation for
          its tagging decision as shown by this heat map of {\em
            learned} attention matrix $A$ for a product
          title. Each map element highlights importance (indicated by
          light color) of a word with respect to neighboring context.}
		\label{fig:heatmap} 
	\end{center}
\vspace{-1em}
\end{figure}

\noindent {\bf Interpretable explanation using attention.}
Figure~\ref{fig:heatmap} shows the heat map of the attention matrix
$A$ --- as learned by \sysname during training --- of a product
title. Each element of the heat map highlights the importance
(represented by a lighter color) of a word with respect to its
neighboring context, and, therefore how it affects the tagging
decision. To give an example, consider the four white boxes located in the center of the figure. They demonstrate that
the two corresponding words ``with'' and ``and'' (in columns) are
important for deciding the tags of the tokens ``beef'' and ``liver''
(in rows) which are potential values for the target {\em
  flavor}-attribute. It makes sense since these are conjunctions
connecting two neighboring {\em flavor} segments. This concrete example shows that
our model has learned the semantics of conjunctions and their
importance for attribute value extraction. This is interesting since
we do not use any part-of-speech tag, parsing, or rule-based annotation as
commonly used in NER tasks.\smallskip


\begin{figure}
	\begin{center}
		\includegraphics[width=1\linewidth]{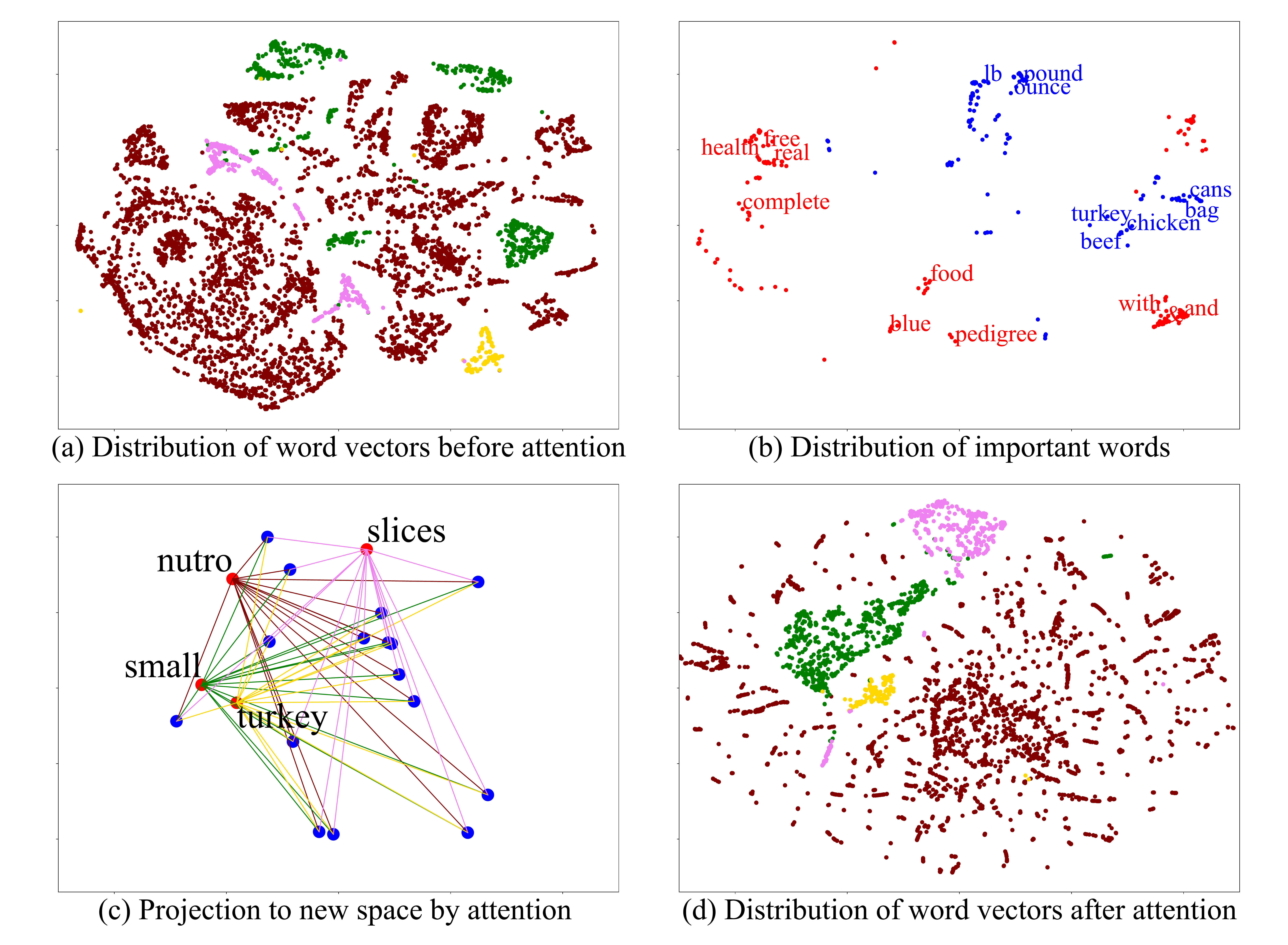}
				\vspace{-6mm}
                \caption{\small Sub-figures (in order) show how \sysname
                  uses {\em attention} to cluster concepts and tags
                  similar to each other in embedding space. Color
                  scheme for tags. $B$: Green, $I$: Gold, $O$: Maroon,
                  $E$: Violet.}
              \vspace{-1em}
		\label{fig:boundary} 
	\end{center}
\end{figure}

\noindent{\bf \sysname achieves better concept
  clustering.} 
The following discussions refer to Figure~\ref{fig:boundary}. The sub-figures in order show how attention operates on the input data to generate better concept clustering. The
corresponding experiments were performed on extracting flavors from dog food titles. For
visualizing high-dimensional word embeddings in a $2$-d plane, we use
t-SNE to reduce the dimensionality of the BiLSTM hidden vector (of
size $200$) to
$2$. 

Figure~\ref{fig:boundary} (a) shows the distribution of word
embeddings {\em before} they are operated by {\em attention} --- where
each dot represents a word token and its color represents a tag
($\{B,I,O,E\}$). We use four different colors to distinguish the
tags. We observe that words with different tags are initially spread
out in the embedding space.

We calculate two importance measures for each word by aggregating corresponding attention weights: (i) its importance
to the {\em attribute words} (that assume any of the $\{B,I,E\}$ tags
for tokens located within an attribute-value), and (ii) its importance
to the {\em outer words} (that assume the tag $O$ for tokens located
outside the attribute-value). For each measure, we sample top $200$
important words and plot them in Figure~\ref{fig:boundary} (b). We
observe that all semantically related words are located close to each
other. For instance, conjunctions like ``with'', ``and'' and ``\&'' are
located together in the right bottom; whereas quantifiers like
``pound'', ``ounce'' and ``lb'' are located together in the top.

We observe that the red dots --- representing the most important words
to the {\em attribute words} --- are placed at the boundary of the
embedding space by the attention mechanism. It indicates that the
attention mechanism is smart enough to make use of the most distinguishable words located in the boundary.  On the other hand, the blue dots ---
denoting the most important words to the {\em outer words} --- are
clustered within the vector space. We observe that quantifiers like
``pound", ``ounce" and ``lb" help to locate the outer words. It is
also interesting that attribute words like ``turkey", ``chicken" and
``beef'' that are extracted as values of the {\em flavor}-attribute
assume an important role in tagging outer words, where these tokens typically signal the boundary of the attribute.


Figure~\ref{fig:boundary} (c) shows how attention mechanism projects
the hidden vectors into a new space. Consider the following example:
``nutro natural choice small breed turkey slices canned dog food, 3.5
oz. by nutro", where ``small breed turkey slices" is the extracted
value of the {\em flavor}-attribute.  Each blue dot in the figure
represents a word of the example in the original hidden space. Red
dots denote the word being projected into a new space by the attention
mechanism. Again, we observe that similar concepts (red dots corresponding to four sample words) come closer to each other {\em after} projection. 

Figure~\ref{fig:boundary} (d) shows the distribution of word vectors
{\em after} being operated by the attention mechanism. Comparing
this with Figure~\ref{fig:boundary} (a), we observe that similar
concepts (tags) now show a better grouping and separability from
different ones after using attention.

\subsection{\sysname with Active Learning: Results}

\subsubsection{Active Learning with Held-Out Test Set}
\begin{figure}[!t]
	\begin{center}
		\includegraphics[width=1.0\linewidth]{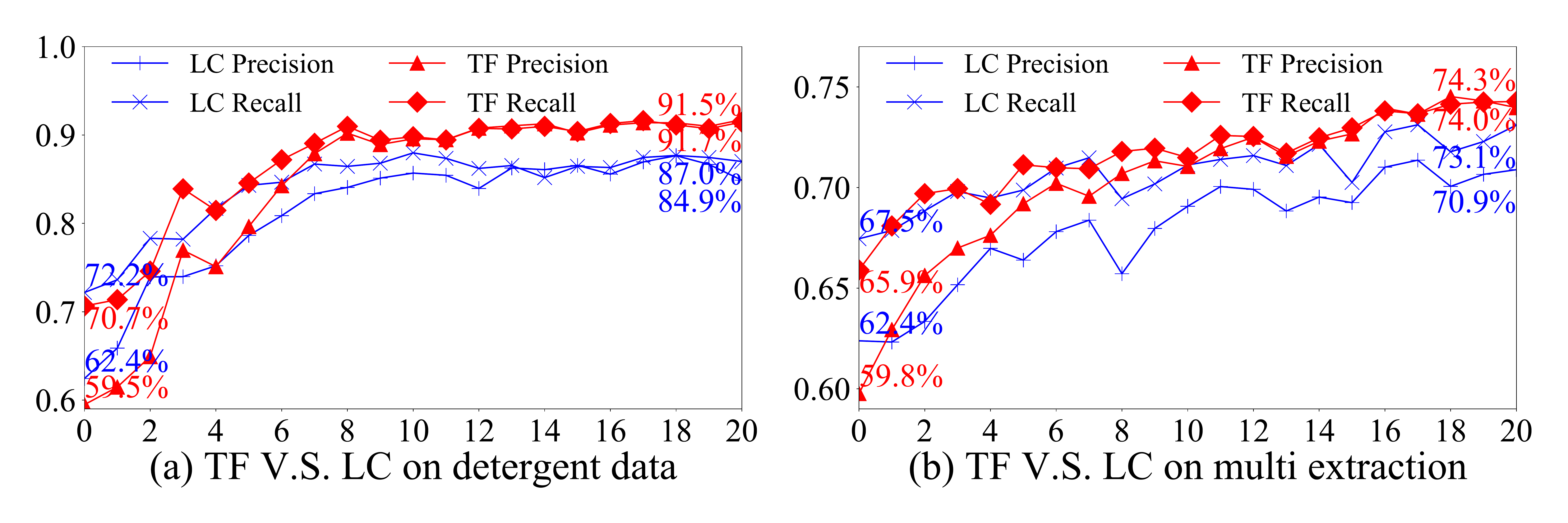}
				\vspace{-7mm}
                \caption{\small \sysname active learning results on
                  held-out test set. \sysname ~{\em with} tag flip (TF)
                  outperforms least confidence (LC)~\cite{Culotta2005}
                  strategy, as well as \sysname ~{\em without} active
                  learning. X-axis shows the number of epochs; Y-axis shows  corresponding precision and recall values. }
		\label{fig:ac_exp3} 
				\vspace{-3mm}
	\end{center}
\end{figure}

In order to have a strict evaluation for the active learning
framework: we use a blind held-out test set $H$ that \sysname cannot
access during training. The original test set in
Table~\ref{tab:dataset} is randomly split into unlabeled pool $U$ and
held-out test set $H$ with the ratio of $2:1$. We start with a very
small number of labeled instances, namely $50$ randomly sampled
instances, as our initial labeled set $L$. We employ $20$ rounds of
active learning for this experiment. Figure~\ref{fig:ac_exp3} shows
the results on two tasks: (i) extracting values for {\em
  scent}-attribute from titles of detergent products, and (ii)
extracting values for multiple attributes {\em brand, capacity} and
{\em flavor} from titles of dog food products.

\sysname with tag flip sampling strategy for single attribute value
extraction improves the precision from $59.5\%$ (on our initial
labeled set of $50$ instances) to $91.7\%$ and recall from $70.7\%$ to
$91.5\%$. This is also better than the results reported in
Table~\ref{tab:overall-perf}}, where \sysname obtained $84.5\%$
precision and $88.2\%$ recall -- trained on the entire data
set. Similar results hold true for multi-attribute value extraction.

We also observe that the tag flip strategy (TF) outperforms least
confidence (LC)~\cite{Culotta2005} strategy by $5.6\%$ in f-score for
single attribute value extraction and by $2.2\%$ for multi-attribute
value extraction.


\subsubsection{Active Learning without Held-out Data}

\begin{figure}[!t]
	\begin{center}
		\includegraphics[width=1.0\linewidth]{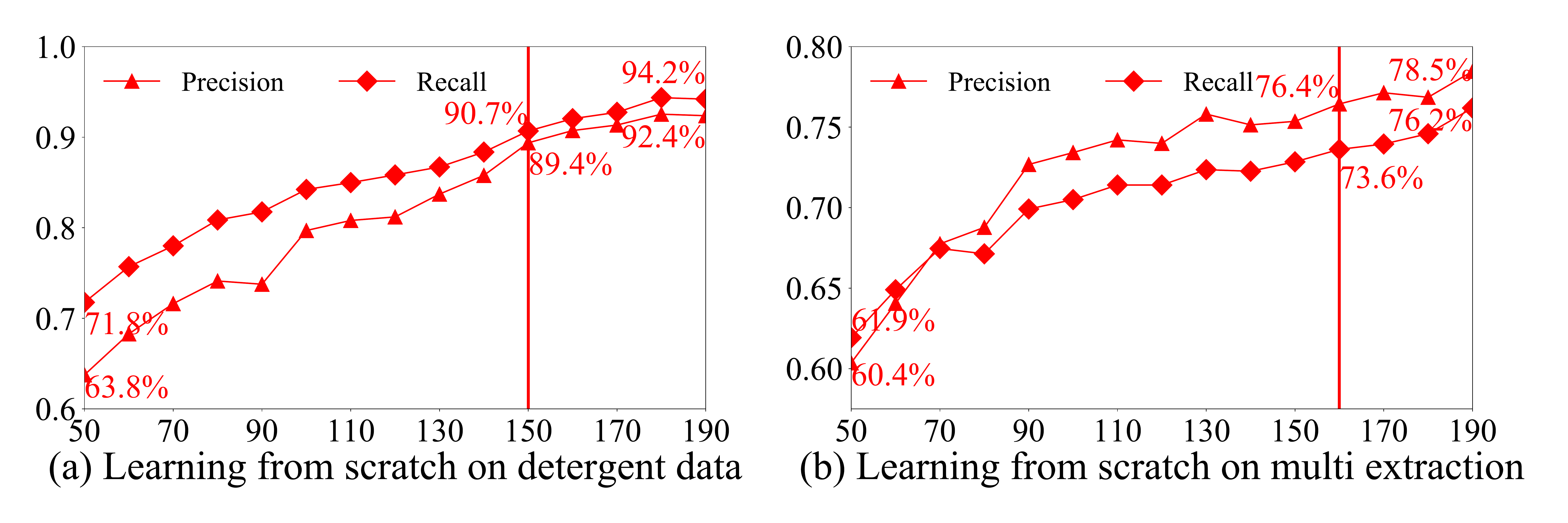}
				\vspace{-6mm}
                \caption{\small Results of active learning with
                  tag flip. \sysname reduces burden of human annotation
                  by $3.3x$. X-axis shows number of annotations; Y-axis shows  corresponding precision and recall values. }
		\label{fig:ac_exp2} \vspace{-4mm}
	\end{center}
\end{figure}

Next we explore to what extent can active learning reduce the burden
of human annotation. As before, we start with a very small number
($50$) of labeled instances as initial training set $L$. We want
to find: how many rounds of active learning are required to match
the performance of \sysname as in original training data of $500$
labeled instances. In contrast to the previous setting with a held-out test
set, in this experiment \sysname can access all of the unlabeled data
to query for their labels. Figure~\ref{fig:ac_exp2} shows its results.
 
For this setting, we use the best performing query strategy from the
last section using tag flips (TF). We achieve almost the same level of
performance with only $150$ training instances that we had initially
obtained with $500$ training instances in the previous
section. Figure~\ref{fig:ac_exp2} (b) shows a similar result for the
second task as well. This shows that \sysname with TF query
strategy for active learning can drastically cut down on the
requirement of labeled training data.

%% file: related.tex
\section{Related Work}\label{sec:related}
Rule-based extraction techniques~\cite{ner-sekine2007} make use of
domain-specific vocabulary or dictionary to spot key phrases and
attributes. These suffer from limited coverage and closed world
assumptions. Similarly, rule-based and linguistic
approaches~\cite{Chiticariu2010, Mikheev1999} leveraging syntactic
structure of sentences to extract dependency relations do not work
well on irregular structures like titles.

An NER system was built \cite{Putthividhya2011} to annotate brands in
product listings of apparel products. Comparing results of SVM,
MaxEnt, and CRF, they found CRF to perform the best. They used seed
dictionaries containing over $6,000$ known brands for bootstrapping.
A similar NER system was built \cite{DBLP:journals/corr/More16} to tag
brands in product titles leveraging existing brand values. In contrast
to these, we do not use any dictionaries for bootstrapping, and can
discover new values.

There has been quite a few works on applying neural networks for
sequence tagging.  A multi-label multi-class Perceptron classifier for
NER is used by \cite{Ling2012}. They used linear chain CRF to segment
text with BIO tagging. An LSTM-CRF model is used \cite{Kozareva2016}
for product attribute tagging for brands and models with a lot of
hand-crafted features. They used $37,000$ manually labeled search
queries to train their model. In contrast, OpenTag does not use
hand-crafted features, and uses active learning to reduce burden of
annotation.

Early attempts include \cite{petasis2000symbolic, hammerton2003named},
which apply feed-forward neural networks (FFNN) and LSTM to NER
tasks. Collobert et al. \cite{collobert2011natural} combine deep FFNN
and word embedding~\cite{Mikolov2013} to explore many NLP tasks
including POS tagging, chunking and NER. Character-level CNNs were
integrated \cite{santos2015boosting} to augment feature
representation, and their model was later enhanced by LSTM
\cite{chiu2015named}. Huang et al. \cite{Huang2015} adopts CRF with
BiLSTM for jointly modeling sequence tagging.  However they use heavy
feature engineering. Lample et al. \cite{Lample2016} use BiLSTM to
encode both character-level and word-level feature, thus constructing
an end-to-end BiLSTM-CRF solution for sequence tagging. Ma et
al. \cite{ma2016end} replace the character-level model with
CNNs. Currently, BiLSTM-CRF models as above is state-of-the-art
for NER. Unlike prior
works, 
OpenTag uses attention to improve feature representation and 
gives interpretable explanation of its decisions.
Bahdanau et al. \cite{Bahdanau2014} successfully applied attention for
alignment in NMT systems. Similar mechanisms have recently been
applied in other NLP tasks like machine reading and parsing
\cite{cheng2016long,vaswani2017attention}.

Early active learning for sequence labeling research
\cite{scheffer2001active,culotta2005reducing} employ least
confidence (LC) sampling strategies.  Settles and Craven made a
thorough analysis over other strategies and proposed their entropy based
strategies in \cite{Settles2008}.  However, the sampling strategy of
OpenTag is different from them.

%% file: conclusion.tex
\section{Conclusions}\label{sec:conclusion}

We presented OpenTag --- an end-to-end tagging model leveraging
BiLSTM, CRF and Attention --- for imputation of missing attribute
values from product profile information like titles, descriptions and
bullets. OpenTag does not use any dictionary or hand-crafted features
for learning. It also does not make any assumption about the structure of the input data, and, therefore, could be applied to any kind of textual data. The other advantages of OpenTag are: (1) {\em Open World
  Assumption (OWA):} It can discover new attribute values (e.g.,
emerging brands) that it has never seen before, as well as multi-word
attribute values and multiple attributes.  (2) {\em Irregular
  structure and sparse context:} It can handle unstructured text like
profile information that lacks regular grammatical structure with
stacking of several attributes, and a sparse
context. 
(3) {\em Limited annotated data:} Unlike other supervised models and
neural networks, OpenTag requires less training data. It exploits {\em
  active learning} to reduce the burden of human annotation.  (4) {\em
  Interpretability:} OpenTag exploits an {\em attention} mechanism to
generate explanations for its verdicts that makes it easier to debug. We presented experiments in real-life datasets in
different domains where OpenTag discovers {\em new} attribute values
from as few as $150$ annotated samples (reduction in $3.3$x amount of
annotation effort) with a high F-score of $83\%$, outperforming
state-of-the-art models.

\section*{Acknowledgments}
Guineng Zheng and Feifei Li were partially supported by NSF grants 1443046 and 1619287. Feifei Li was also supported in part by NSFC grant 61428204.
The authors would like to sincerely thank Christos Faloutsos, Kevin Small and Ron Benson for their insightful and constructive comments on the paper, and Arash Einolghozati for his contributions to the project. We would also like to thank the following students for helping with annotation: Min Du, Neeka Ebrahimi, Richie Frost, Dyllon Gagnier, Debjyoti Paul, Yuwei Wang, Zhuoyue Zhao, Dong Xie, Yanqing Peng, Ya Gao, Madeline MacDonald.